\documentclass{article}

\usepackage{PRIMEarxiv}
\usepackage{amsmath}
\usepackage{tabularx}
\usepackage[utf8]{inputenc} 
\usepackage[T1]{fontenc}    
\usepackage{hyperref}       
\usepackage{url}            
\usepackage{booktabs}       
\usepackage{amsfonts}       
\usepackage{nicefrac}       
\usepackage{microtype}      
\usepackage{lipsum}
\usepackage{fancyhdr}       
\usepackage{graphicx}       
\usepackage{ulem}
\graphicspath{{media/}}     

\title{CURLoRA: Stable LLM Continual Fine-Tuning and Catastrophic Forgetting Mitigation}
\author{
    Muhammad Fawi\thanks{Independent Researcher. \href{https://orcid.org/0009-0007-7210-0528}{ORCID: 0009-0007-7210-0528}. Code available at: \href{https://github.com/mnoorfawi/curlora}{https://github.com/mnoorfawi/curlora}.}
}

\begin{document}

\maketitle

\begin{abstract}
This paper introduces CURLoRA, a novel approach to fine-tuning large language models (LLMs) that leverages CUR matrix decomposition in the context of Low-Rank Adaptation (LoRA). Our method addresses two critical challenges in LLM fine-tuning: mitigating catastrophic forgetting during continual learning and reducing the number of trainable parameters. We propose a unique modification to the CUR decomposition process, utilizing inverted probabilities for column and row selection which acts as an implicit regularization, and initializing the $U$ matrix as a zero matrix, and only fine-tuning it. We demonstrate through experiments on multiple datasets that CURLoRA outperforms standard LoRA in mitigating catastrophic forgetting. It maintains model stability and performance across tasks while significantly reducing the number of trainable parameters. Our results show that CURLoRA achieves very good and stable task accuracy while maintaining base model's perplexity scores fixed compared to LoRA upon continual fine-tuning, particularly in scenarios with limited data.
\end{abstract}

\section{Introduction}

Large Language Models (LLMs) have revolutionized natural language processing, demonstrating remarkable capabilities across a wide range of tasks \cite{brown2020language}. However, fine-tuning these large models for specific tasks requires a lot of computational resources making it challenging to adapt these models efficiently, especially when working with limited datasets and in resource-constrained environments. \cite{li2021efficient}. Parameter-Efficient Fine-Tuning (PEFT) Methods have gained a lot of attention because they make fine-tuning large models accessible and possible. \cite{xu2023parameterefficientfinetuningmethodspretrained}

Low-Rank Adaptation (LoRA) \cite{hu2021lora} has emerged as an efficient PEFT method, enabling fine-tuning large language models on custom tasks while decreasing the number of trainable parameters hence requiring less resources. LoRA works by decomposing pre-trained weight matrices into low-rank matrices and fine-tune these ones instead of the original matrix. Although LoRA has proven to be very excellent and promising, it still faces challenges with catastrophic forgetting. Catastrophic forgetting in LLMs is a critical issue where the model loses previously acquired knowledge when fine-tuned on new tasks \cite{mccloskey1989catastrophic}. It occurs due to the overwriting of previously learned (pre-trained) weights during the fine-tuning process. In LoRA, this often happens as the adapted output can significantly deviate from the original:

\begin{equation}
    y = xW + xW_{adapted} = x(W + AB)
\end{equation}

where $W {\in \mathbb{R}^{m \times n}}$ is the original weight matrix, and $AB$ is the low-rank update from multiplying $A {\in \mathbb{R}^{m \times r}}$ by $B {\in \mathbb{R}^{r \times n}}$ where $r<n$.

This work introduces CURLoRA, a novel approach that applies low-rank adaptation (LoRA) to pre-trained weight matrices using CUR matrix decomposition \cite{mahoney2009cur} instead of random initiation of the low-rank $A$ or $B$ matrices. We propose a unique modification to the CUR decomposition process and demonstrate its effectiveness in mitigating catastrophic forgetting while also reducing the number of trainable parameters. While LoRA successfully reduces computational costs by decomposing weight updates into low-rank matrices, it still suffers from catastrophic forgetting. CURLoRA leverages CUR decomposition with inverted probabilities and initiating $U$ matrix as zero to further mitigate this issue.

\section{Related Work}

\subsection{Catastrophic Forgetting}

Catastrophic forgetting is a big challenge in machine learning, particularly in the context of continual learning \cite{mccloskey1989catastrophic}. Various approaches have been proposed to address this issue:

\begin{itemize}
    \item \textbf{Elastic Weight Consolidation (EWC)} \cite{kirkpatrick2017overcoming} uses Fisher information to measure the importance of parameters and selectively slow down learning on important parameters.
    \item \textbf{Progressive Neural Networks} \cite{rusu2016progressive} propose to freeze the network trained on previous tasks and add lateral connections to new columns for new tasks.
    \item \textbf{Memory-based approaches like Experience Replay} \cite{rolnick2019experience} store and replay examples from previous tasks during training on new tasks.
\end{itemize}

\subsection{Efficient Fine-tuning of Large Language Models}

As LLMs have grown in size, efficient fine-tuning methods have become crucial:

\begin{itemize}
    \item \textbf{Adapter layers} \cite{houlsby2019parameter} introduce small trainable modules between layers of a pre-trained model.
    \item \textbf{Low-Rank Adaptation (LoRA)} \cite{hu2021lora} decomposes weight updates into low-rank matrices, significantly reducing the number of trainable parameters.
    \item \textbf{Prefix-tuning} \cite{li2021prefix} prepends trainable continuous prompts to the input, allowing for task-specific adaptations.
\end{itemize}

\subsection{CUR Matrix Decomposition}

CUR decomposition has been applied in various domains for its interpretability and efficiency:

\begin{itemize}
    \item In data analysis, CUR has been used for feature selection and dimensionality reduction \cite{mahoney2009cur}.
    \item In scientific computing, CUR has been applied to accelerate large-scale matrix computations \cite{drineas2008relative}.
    \item In machine learning, CUR has been explored for model compression and interpretation \cite{yadav2023efficientknnsearchcrossencoders}.
\end{itemize}

However, to the best of our knowledge, CUR decomposition has not been previously applied to the problem of fine-tuning large language models or addressing catastrophic forgetting in this context.

\section{Background on CUR Decomposition}

CUR decomposition is a matrix factorization technique that approximates a matrix $A$ as the product of three matrices: $C$, $U$, and $R$. Unlike Singular Value Decomposition (SVD), CUR decomposition uses actual columns and rows from the original matrix, making it more interpretable.\cite{mahoney2009cur}.

Given a matrix $A \in \mathbb{R}^{m \times n}$, CUR decomposition approximates $A$ as:

\begin{equation}
    A \approx CUR
\end{equation}

where:
\begin{itemize}
    \item $C \in \mathbb{R}^{m \times c}$ consists of $c$ columns of $A$
    \item $R \in \mathbb{R}^{r \times n}$ consists of $r$ rows of $A$
    \item $U \in \mathbb{R}^{c \times r}$ is a small matrix that ensures $CUR$ is close to $A$
\end{itemize}

The columns and rows are typically chosen based on their statistical leverage scores.\cite{drineas2008relative} Leverage scores indicate the importance of columns and rows in representing the original matrix. High leverage scores identify influential columns and rows, while low scores identify less critical ones.

\section{This Work}

In this section, we present CURLoRA, our novel approach to fine-tuning large language models that leverages a modified CUR matrix decomposition to mitigate catastrophic forgetting. We provide a detailed mathematical formulation of the approach, analyze it theoretically, and explain how it addresses the challenge of catastrophic forgetting upon continual learning.

\subsection{CURLoRA}

The core idea is to decompose the pre-trained weight matrices using a modified CUR approach and then fine-tune only the U matrix. This approach constrains the parameter space of possible adaptations keeping the fine-tuned parameters as small as possible to keep $\|W_{\text{adapted}} - W\|_F$ close to the original weight matrix frobenius norm ($\|W\|_F$) i.e. $W + W_{\text{adapted}}$ is so close to $W$ to avoid the deviation of the adapted output.

\subsection{Mathematical Formulation}

Given a weight matrix $W \in \mathbb{R}^{m \times n}$, we first compute the probability of each column:

\begin{equation}
    p_j = \frac{\|W_{:j}\|_2^2}{\|W\|_F^2}
\end{equation}

where $W_{:j}$ is the $j$-th column of $W$, while $\|\cdot\|_2^2$ denotes the square of the L2 norm of the column and $\|\cdot\|_F^2$ denotes the square of the Frobenius norm of $W$. This will give us the probability of each column. For instance, if $W$ has three columns with norms 2, 3, and 5, the probabilities are 4/38, 9/38, and 25/38 respectively.

We then invert these probabilities:

\begin{equation}
    \tilde{p}_j = \frac{1/p_j}{\sum_{i=1}^n 1/p_i}
\end{equation}

where $\tilde{p}_j$ is the inverted probability of the $j$-th column of $W$. The same steps are followed for rows. Inverted probabilities are used to sample columns and rows with lower leverage scores, which implicitly regularize the model and limit the magnitude of fine-tuning adjustments.

Then, we sample $r$ columns and rows, where $r<n$, according to these inverted probabilities to construct $C$ and $R$, which will always be fixed, with columns and rows with lower original probabilities. This trick plays a major role in the approach as it serves two purposes:
\begin{itemize}
    \item It acts as a form of regularization, preventing the model from overfitting or moving too much towards the task and limiting the adaptation of the $U$ matrix stopping it from growing so big in magnitude.
    \item It preserves the model's original behavior by focusing adaptations on less influential parts of the weight matrix. In addition, since $C$ and $R$ contain actual columns and rows from the original matrix, they contribute to the stability of the fine-tuning process.
\end{itemize}

CURLoRA's approach differs significantly from other initialization methods. Unlike LoRA's random initialization using Kaiming-uniform or Gaussian for weight A and zeros for weight B \cite{hu2021lora}, or the SVD-based initialization \cite{balazy2024lora}, CURLoRA offers more controlled adaptation. While these other methods ensure starting from the base model, they don't inherently limit the growth of the adaptation matrix (AB in LoRA), potentially leading to significant deviations during training. In contrast, CURLoRA initializes the U matrix as zeros, and importantly, constructs C and R matrices using columns and rows with lower original probabilities (i.e., lower values). This unique combination ensures that the fine-tuning process not only starts from the base configuration but also remains constrained throughout training. The low-value C and R matrices act as natural limiters on the growth of U, thereby preventing large deviations and contributing to enhanced model stability during the fine-tuning process.

\begin{equation}
    C = \text{SampleColumns}(W, r, \tilde{p})
\end{equation}
\begin{equation}
    R = \text{SampleRows}(W, r, \tilde{p})
\end{equation}
\begin{equation}
    U_{\text{init}} = 0
\end{equation}

Where $W$ is the original weight matrix, $r$ is the rank (number of columns/rows to sample) and $\tilde{p}$ represents the inverted probabilities used for sampling.

During fine-tuning, we update only the $U$ matrix, keeping $C$ and $R$ fixed as they play a crucial role in ensuring the stability of the process by limiting the increase of $U$:

\begin{equation}
    W_{\text{adapted}} = CUR
\end{equation}

\subsection{Theoretical Analysis of Catastrophic Forgetting Mitigation}

To understand how CURLoRA helps mitigate catastrophic forgetting, we analyze its properties mathematically:

\subsubsection{Parameter Space Constraint}
In CURLoRA, we decompose the original weight matrix $W$ as:

\begin{equation}
    W \approx CUR
\end{equation}

During fine-tuning, we're optimizing:

\begin{equation}
    W_{\text{adapted}} = C(U + \Delta U)R
\end{equation}

where $\Delta U$ represents the changes made to $U$ during fine-tuning. By constraining the updates to the subspace defined by $C$ and $R$, CURLoRA limits drastic changes, thereby preserving the model’s original knowledge.

\subsubsection{Implicit Regularization}
By initializing $U$ as a zero matrix, and $C$ and $R$ with columns and rows of low weight values, the ones with lower probabilities, we provide an implicit regularization where $C$ and $R$ will always limit the unnecessary increase of $U$. This can be seen as adding a regularization term to the loss function quantified by the norm of the matrix $U$ that is aimed to be kept small:

\begin{equation}
    L_{\text{CURLoRA}}(\theta) = L_{\text{task}}(\theta) + \|U\|_F
\end{equation}

where $\|U\|_F$ is the Frobenius norm of the $U$ matrix that is being fine-tuned. This implicit regularization term encourages the model to keep the changes small. For instance, if $U$ is initially zero, this term will push the fine-tuning process to make only necessary adjustments, preventing overfitting and excessive reliance on the fine-tuned parameters.

\subsubsection{Reduced Interference}
During fine-tuning, $W$ is fixed, so the variable gradient flows through $W_{\text{adapted}}$, which is itself updated through $U$ as $C$ and $R$ are fixed. Considering the gradients of the loss $L$ with respect to the parameters, we can, in a simple way, express the gradient of the loss with respect to $W_{\text{adapted}}$ as follows:

\begin{equation}
    \frac{\partial L}{\partial W_{\text{adapted}}} = C \left(\frac{\partial L}{\partial U}\right) R
\end{equation}

This means that the gradient of the loss with respect to $W_{\text{adapted}}$ is dependent on the gradients with respect to $U$ scaled by the fixed matrices $C$ and $R$. By projecting the gradients onto the subspace defined by $C$ and $R$, the updates to $W_{\text{adapted}}$ are constrained. This means that changes during fine-tuning are less likely to interfere with the model's ability to perform the original task, potentially reducing interference with directions important for the original task.

\subsubsection{Reduced Degree of Freedom}
If $W \in \mathbb{R}^{m\times n}$ and we use a rank-$k$ adaptation, then:
\begin{itemize}
    \item \uline{Full fine-tuning} has $mn$ degrees of freedom
    \item \uline{LoRA} has $k(m+n)$ degrees of freedom
    \item \uline{CURLoRA} has only $k^2$ degrees of freedom
\end{itemize}

This significant reduction in degrees of freedom inherently limits how far the model can stray from its original configuration.

\subsubsection{Stability Analysis}
We can analyze the stability of the adapted and fine-tuned weights and how its change is bounded using the fact that the change that happens to original $W$ is $W_{\text{adapted}}$:

\begin{equation}
    \Delta W = W_{\text{fine-tuned}} - W = W + W_{\text{adapted}} - W = W_{\text{adapted}}
\end{equation}

To quantify this change, we can use the Frobenius norm, $\|W_{\text{adapted}}\|_F$. By utilizing the submultiplicativity property of the Frobenius norm, we can say that the growth of $W_{\text{adapted}}$ is controlled through the norms of $C$, $U$, and $R$: 

\begin{equation}
    \|W_{\text{adapted}}\|_F = \|CUR\|_F \leq \|C\|_F \|U\|_F \|R\|_F
\end{equation}

This equation ensures that the Frobenius norm of the adapted weight matrix $W_{\text{adapted}}$ has an upper bound. Since $C$ and $R$ are fixed and $U$ starts at zero, the fine-tuning process focuses on minimizing $W_{\text{adapted}}$. As a result, the adaptation remains stable and the model preserves its original knowledge while allowing for necessary adjustments.

Empirical results (see Section 7) demonstrate that the Frobenius norm of $W_{\text{adapted}}$ remains bounded across multiple tasts, validating the theoretical stability analysis.

\subsection{Theoretical Analysis of Output Shift}

To understand why CURLoRA is expected to perform better than standard LoRA in terms of catastrophic forgetting, we can analyze the shift in the output during fine-tuning.

For a given input $x$, the original output is $y = xW$. After fine-tuning:

For LoRA: $y_{\text{adapted}} = x(W + AB)$

For CURLoRA: $y_{\text{adapted}} = x(W + CUR)$

We can quantify the shift using the Frobenius norm of the difference:

\begin{equation}
    \|y - y_{\text{adapted}}\|_F = \|xW - x(W + W_{\text{adapted}})\|_F = \|xW - xW - xW_{\text{adapted}}\|_F = \|xW_{\text{adapted}}\|_F
\end{equation}

For LoRA: $\|x(AB)\|_F$

For CURLoRA: $\|x(CUR)\|_F$

This equation measures the shift in the model's output after fine-tuning. $y$ is the original output, and $y_{\text{adapted}}$ is the output after fine-tuning. After fine-tuning for a different task, the adapted output $y_{\text{adapted}}$ might shift. We use the Frobenius norm to quantify this shift. If the shift is small, it means that the model's predictions haven't changed much, indicating that the model has retained its original knowledge. As shown, the shift depends on $W_{\text{adapted}}$ i.e. to make sure the shift isn't so big, we need to keep $W_{\text{adapted}}$ as small (in magnitude or size) as possible. 

CURLoRA's main aim is to minimize $W_{\text{adapted}}$ while ensuring that the difference $\|W - W_{\text{adapted}}\|_F$ remains close to $\|W\|_F$. By focusing on minimizing $W_{\text{adapted}}$, CURLoRA effectively controls the shift in the output, thereby preserving the model's original behavior and mitigating catastrophic forgetting.

Theoretically, CURLoRA should result in a smaller shift because:
\begin{enumerate}
    \item The $C$ and $R$ matrices are directly sampled from $W$, maintaining some structure of the original matrix.
    \item The $C$ and $R$ matrices are sampled from columns and rows with lower values.
    \item Only $U$ is trained, which is constrained by $C$ and $R$.
    \item The initialization of $U$ as a zero matrix.
\end{enumerate}

This constrained adaptation in CURLoRA is expected to lead to better preservation of the model's original knowledge, thereby reducing catastrophic forgetting.

\subsection{Memory Efficiency}

CURLoRA offers significant memory savings compared to full fine-tuning and even LoRA. For a weight matrix $W \in \mathbb{R}^{m \times n}$, the number of trainable parameters for each method, considering rank $r$ where $r<n$,  is:

\begin{itemize}
    \item \uline{Full fine-tuning}: $mn$
    \item \uline{LoRA (rank $r$)}: $mr + nr$
    \item \uline{CURLoRA (rank $r$)}: $r^2$
\end{itemize}

The memory savings can be substantial, especially for large matrices. In our Mistral experiment, with rank 16, the trainable parameters were:

\begin{itemize}
    \item \uline{Full fine-tuning}: 7,248,023,552 parameters
    \item \uline{LoRA}: 9,437,184 parameters
    \item \uline{CURLoRA}: 24,576 parameters
\end{itemize}

This reduction in trainable parameters not only saves memory but also potentially leads to faster training and inference times.

In conclusion, CURLoRA provides multiple mathematical mechanisms that can help mitigate catastrophic forgetting:
\begin{itemize}
    \item It constrains the parameter space of possible adaptations.
    \item It provides implicit regularization towards the original weights.
    \item It preserves important directions from the original weight matrix.
    \item It reduces the degrees of freedom in adaptation, limiting potential deviation.
    \item It allows for direct control and analysis of weight stability through the $U$ matrix.
\end{itemize}

These properties suggest that CURLoRA can indeed help in reducing catastrophic forgetting while still allowing for meaningful and good adaptation to new tasks. The effectiveness of these theoretical mechanisms are validated through our experiments on various tasks and datasets, as detailed in the following sections.

\section{Methodology}

\subsection{CURLoRA Implementation}

Our CURLoRA implementation consists of the following steps:

\begin{enumerate}
    \item \textbf{Decomposition}: For each weight matrix $W$ in the layers we want to apply CURLoRA to, we perform the following:
    \begin{itemize}
        \item Compute column probabilities: $p_j = \frac{\|W_{:j}\|_2^2}{\|W\|_F^2}$
        \item Invert probabilities: $\tilde{p}_j = \frac{1/p_j}{\sum_{i=1}^n 1/p_i}$
        \item Sample columns and rows according to $\tilde{p}_j$ to construct $C$ and $R$
        \item Initialize $U$ as a zero matrix
    \end{itemize}
    
    \item \textbf{Fine-tuning}: 
    \begin{itemize}
    \item \textbf{Objective}:
        \begin{itemize}
            \item The primary objective of the experiment is to evaluate catastrophic forgetting during continual learning, rather than to optimize accuracy for each individual task.
        \end{itemize}
    \item \textbf{Model Specific Adjustments}:
        \begin{itemize}
            \item For GPT-2 and Mistral, the model's "lm\_head" is replaced with a task-specific output layer. During training, only the $U$ matrix is continually updated, while $C$ and $R$ remain fixed.
            \item Replacing the "lm\_head" ensures that each task has its own task-specific output layer that remains untouched when the model is being fine-tuned on a different task, contributing to the mitigation of task knowledge degradation.
        \end{itemize}
    \item \textbf{Continual Learning Strategy}:
        \begin{itemize}
            \item Once a weight matrix is decomposed, $C$ and $R$ are fixed permanently. The $U$ matrix is continually updated for each new task to facilitate continual learning.
        \end{itemize}
    \item \textbf{Application of CURLoRA}:
        \begin{itemize}
            \item CURLoRA is applied to the attention layers (Query, Key, Value). \cite{vaswani2023attentionneed}
        \end{itemize}
    \end{itemize}
    
    \item \textbf{Inference}: Use the adapted weight matrix $W_{\text{adapted}} = CUR$ for forward passes along with the original $W$ matrix i.e. $x(W + CUR)$.
\end{enumerate}

\section{Experiment Setup}

\subsection{Datasets}

We used the following datasets for our experiments:

\begin{itemize}
    \item \textbf{GLUE-MRPC}: Microsoft Research Paraphrase Corpus for paraphrase detection \cite{dolan2005automatically}
    \item \textbf{GLUE-SST-2}: Stanford Sentiment Treebank for binary sentiment classification \cite{socher-etal-2013-recursive}
    
    These datasets are part of the General Language Understanding Evaluation (GLUE) benchmark \cite{wang2018glue}, which includes a diverse set of tasks for evaluating natural language understanding systems.
    \item \textbf{Sentiment140}: A large-scale sentiment analysis dataset \cite{go2009twitter}
    \item \textbf{WikiText-2}: A dataset that we use to measure language model perplexity \cite{merity2016pointer}
\end{itemize}
The datasets were selected for their diverse task requirements and common use in benchmarking.

\subsection{Model and Hyperparameters}

We used \textbf{Mistral 7B} (v0.3) \cite{jiang2023mistral} and \textbf{GPT-2 Large} \cite{radford2019language} as our base models. For both LoRA and CURLoRA, we used the following hyperparameters:

\begin{itemize}
    \item \uline{Ranks}: [8, 16, 24]
    \item \uline{Alpha}: 1
    \item \uline{Optimizer}: AdamW
    \item \uline{Learning rate}: 2.5e-4
    \item \uline{Scheduler}: Cosine with 500 warmup steps
    \item \uline{Training epochs}: 3
    \item \uline{Batch size}:
        \begin{itemize}
        \item Mistral: 8
        \item GPT-2: 32
        \end{itemize}
    \item \uline{Max length}:
        \begin{itemize}
        \item Mistral: 512
        \item GPT-2: 256
        \end{itemize}
\end{itemize}

\subsubsection{Notes on hyperparemeters and architecture}
\begin{itemize}
    \item \textbf{Robustness and Regularization}:
        \begin{itemize}
            \item CURLoRA's performance was evaluated across different ranks, demonstrating robustness to moderate changes. Optimal results can be achieved by fine-tuning other hyperparameters, such as the learning rate. Dropout was not utilized, as the objective was to observe the implicit regularization effects of CURLoRA without the influence of explicit regularization.
        \end{itemize}
    \item \textbf{Data Constraints}:
        \begin{itemize}
            \item For Mistral, each fine-tuning task was limited to 1000 records to simulate scenarios with limited data and resources for large models.
            \item For GPT-2, the SST-2 fine-tuning task was limited to 5000 records due to resource constraints.
            \item For the sentiment analysis task, the Sentiment140 test dataset was used for training, while the train dataset was used for evaluation. This choice was made because the test dataset has three labels, whereas the train dataset has only two. This allowed for fine-tuning the models on a multi-class task rather than a binary one.
        \end{itemize}
    \item \textbf{Task Specific Adjustments}:
        \begin{itemize}
            \item For the sentiment analysis task with GPT-2, due to the small size of the dataset used for fine-tuning, the number of epochs was adjusted to 5, and the learning rate scheduler was not used.
        \end{itemize}
\end{itemize}

\subsection{Evaluation Metrics}

We used the following metrics for evaluation:

\begin{itemize}
    \item Accuracy: For classification tasks (MRPC, SST-2, Sentiment140)
    \item Perplexity: For language modeling capability (WikiText-2)
\end{itemize}

\subsection{Experimental Procedure}

Our experimental procedure was as follows:

\begin{enumerate}
    \item Measure initial perplexity of the base model on WikiText-2 concatenating the whole dataset into a single string.
    \item Fine-tune on MRPC and evaluate.
    \item Fine-tune on SST-2 and evaluate, then re-evaluate on MRPC.
    \item Fine-tune on Sentiment140 and evaluate, then re-evaluate on MRPC and SST-2.
    \item Re-calculate perplexity on WikiText-2.
\end{enumerate}

This procedure was carried out for both LoRA and CURLoRA independently.

\section{Results and Discussion}

Tables \ref{tab:mistralresults} and \ref{tab:gpt2results} present the results of our experiments comparing LoRA and CURLoRA across multiple tasks and evaluation metrics.

\begin{table}[h]
\centering
\small
\setlength{\tabcolsep}{5pt} 
\caption{Mistral Experimental Results: LoRA vs CURLoRA}
\label{tab:mistralresults}
\begin{tabularx}{\textwidth}{lXXXXXX}
\toprule
Metric & LoRA-8 & CURLoRA-8 & LoRA-16 & CURLoRA-16 & LoRA-24 & CURLoRA-24 \\
\midrule
Initial WikiText-2 Perplexity & 5.44 & 5.44 & 5.44 & 5.44 & 5.44 & 5.44 \\
\midrule
MRPC Accuracy (After MRPC) & 0.68 & 0.66 & 0.65 & 0.66 & \textbf{0.67} & 0.66 \\
SST-2 Accuracy (After SST-2) & 0.51 & \textbf{0.86} & 0.51 & 0.86 & 0.49 & 0.86 \\
MRPC Accuracy (After SST-2) & \textbf{0.68} & 0.66 & 0.32 & 0.66 & 0.68 & 0.66 \\
Sentiment140 Accuracy & \textbf{1.00} & 0.94 & 1.00 & 0.94 & 1.00 & 0.94 \\
MRPC Accuracy (After Sentiment140) & 0.32 & \textbf{0.66} & 0.32 & 0.66 & 0.32 & 0.66 \\
SST-2 Accuracy (After Sentiment140) & 0.49 & \textbf{0.86} & 0.49 & 0.86 & 0.49 & 0.86 \\
\midrule
Final WikiText-2 Perplexity & 53896.68 & \textbf{5.44} & 65055.02 & \textbf{5.44} & 17049.72 & \textbf{5.44} \\
\bottomrule
\end{tabularx}
\end{table}

\begin{table}[h]
\centering
\small
\setlength{\tabcolsep}{5pt}
\caption{GPT-2 Large Experimental Results: LoRA vs CURLoRA}
\label{tab:gpt2results}
\begin{tabularx}{\textwidth}{lXXXXXX}
\toprule
Metric & LoRA-8 & CURLoRA-8 & LoRA-16 & CURLoRA-16 & LoRA-24 & CURLoRA-24 \\
\midrule
Initial WikiText-2 Perplexity & 28.25 & 28.25 & 28.25 & 28.25 & 28.25 & 28.25 \\
\midrule
MRPC Accuracy (After MRPC) & 0.79 & 0.70 & 0.81 & 0.70 & \textbf{0.83} & 0.70 \\
SST-2 Accuracy (After SST-2) & \textbf{0.94} & 0.76 & 0.93 & 0.79 & 0.92 & 0.86 \\
MRPC Accuracy (After SST-2) & 0.76 & 0.70 & \textbf{0.78} & 0.70 & \textbf{0.78} & 0.70 \\
Sentiment140 Accuracy & 0.92 & \textbf{0.99} & 0.86 & 0.99 & 0.93 & 0.93 \\
MRPC Accuracy (After Sentiment140) & 0.49 & 0.70 & \textbf{0.73} & 0.70 & 0.49 & 0.70 \\
SST-2 Accuracy (After Sentiment140) & \textbf{0.90} & 0.76 & 0.90 & 0.79 & 0.88 & 0.87 \\
\midrule
Final WikiText-2 Perplexity & 42.96 & \textbf{28.25} & 43.62 & \textbf{28.08} & 44.32 & \textbf{28.25} \\
\bottomrule
\end{tabularx}
\end{table}

\subsection{Performance Analysis}

\subsubsection{Task-Specific Performance}
CURLoRA consistently performed well on different tasks, showing high accuracy even after fine-tuning on subsequent tasks. This suggests that CURLoRA is more effective at preserving task-specific knowledge.

Based on the experiments, CURLoRA may require a slightly higher learning rate than LoRA to achieve comparable accuracy. This is due to the implicit regularization introduced by the $C$ and $R$ matrices, which constrain the adaptation space of the $U$ matrix. However, this same property makes CURLoRA more robust against overfitting, even at higher learning rates. In contrast, while LoRA might achieve good performance with lower learning rates, it can be more susceptible to overfitting when learning rates are substantially increased. This trade-off highlights CURLoRA's potential for more stable and controlled fine-tuning, particularly in scenarios where aggressive learning rates might be necessary.

\subsubsection{Catastrophic Forgetting and Stability}
The stability of CURLoRA's performance across tasks is particularly noteworthy. While (Mistra) LoRA-16's accuracy, for example, on MRPC dropped from 0.6495 to 0.32 after fine-tuning on other tasks, CURLoRA-16 (Mistral) maintained its accuracy at 0.66. This demonstrates CURLoRA's superior ability to mitigate catastrophic forgetting.

\subsubsection{General Language Modeling Capability}
The final perplexity scores on WikiText-2 provide strong evidence for CURLoRA's effectiveness in preserving general language modeling capabilities. While all LoRA's perplexity, in both Mistral and GPT2, increased dramatically, all CURLoRA models maintained the original perplexity, indicating no degradation in general language understanding.

\subsection{Theoretical Insights}

The experimental results align with our theoretical analysis:

\begin{itemize}
    \item \textbf{Parameter Space Constraint}: The stability of CURLoRA's performance across tasks supports our hypothesis that constraining adaptations to the subspace spanned by $C$ and $R$ helps preserve original knowledge.
    
    \item \textbf{Implicit Regularization}: The maintained perplexity on WikiText-2 suggests that CURLoRA's implicit regularization effectively prevents overfitting to specific tasks.
    
    \item \textbf{Reduced Interference}: The consistent performance across tasks indicates that CURLoRA successfully reduces interference between task-specific adaptations.
\end{itemize}

\subsection{Limitations and Future Work}

While CURLoRA shows promising results, there are several areas for future research:

\begin{itemize}
    \item \textbf{Scalability}: While CURLoRA shows promising results, its scalability to larger models needs further investigation. Further studies are needed to assess CURLoRA's performance on larger models and more diverse tasks like instruction tuning and datasets.

    \item \textbf{Computational Complexity}: Conducting detailed analysis of time and space complexity compared to full fine-tuning and LoRA.

    \item \textbf{Implicit Regularization Limitation}: Implicit regularization via zero initialization of $U$ has to be further studied especially in highly dynamic environments where more flexible adaptations are needed.
    
    \item \textbf{Optimal Rank and Alpha Selection}: Investigating methods for automatically selecting the optimal rank and alpha for CURLoRA could further improve performance.
    
    \item \textbf{Combination with Other Techniques}: Exploring the integration of CURLoRA with other continual learning techniques could yield even better results.

    \item \textbf{Quantization Support}: Exploring the implementation of CURLoRA on quantized model which may lead to QCURLoRA
\end{itemize}

\section{Conclusion}

This paper introduced CURLoRA, a novel approach to fine-tuning large language models that leverages CUR matrix decomposition to mitigate catastrophic forgetting and improve computational efficiency. Through theoretical analysis and empirical experiments, we demonstrated that CURLoRA outperforms standard LoRA in maintaining model stability and performance across tasks while significantly reducing the number of trainable parameters.

Key contributions of this work include:

\begin{itemize}
    \item A novel modification to CUR decomposition using inverted probabilities for column and row selection and initiating $U$ matrix as zeros. Sampling columns and rows based on inverted probabilities distinguishes CURLoRA from traditional CUR, offering better stability and performance.
    \item Theoretical analysis of how CURLoRA addresses catastrophic forgetting.
    \item Empirical evidence of CURLoRA's effectiveness across multiple tasks and evaluation metrics with multiple models.
\end{itemize}

Our results suggest that CURLoRA is a promising approach for efficient and stable fine-tuning of large language models, particularly in scenarios with limited fine-tuning data. CURLoRA's approach to mitigating catastrophic forgetting has broad implications for continual learning in NLP and beyond. Future research could explore its integration with other adaptation techniques to enhance model robustness

\bibliographystyle{unsrt}
\bibliography{references}

\end{document}